%% file: ms.tex
\documentclass{article}

\usepackage{microtype}
\usepackage{graphicx}
\usepackage{subfigure}
\usepackage{booktabs} 

\usepackage{hyperref}

\usepackage[accepted]{icml2019}

\usepackage{mathtools}
\usepackage{amssymb}
\usepackage{amsthm}

\newcommand{\KL}[2]{ D \left[ #1 \, || \, #2 \right]}		

\newtheorem{lemma}{Lemma}

\icmltitlerunning{Moment-Based Variational Inference for Markov Jump Processes}

\begin{document}

\twocolumn[
\icmltitle{Moment-Based Variational Inference for Markov Jump Processes }



\icmlsetsymbol{equal}{*}

\begin{icmlauthorlist}
\icmlauthor{Christian Wildner}{to}
\icmlauthor{Heinz Koeppl}{to}
\end{icmlauthorlist}

\icmlaffiliation{to}{Department of Electrical Engineering and Information Technology, Technische Universit\"at Darmstadt, Germany}

\icmlcorrespondingauthor{Heinz Koeppl}{heinz.koeppl@bcs.tu-darmstadt.de}

\icmlkeywords{Machine Learning, ICML}

\vskip 0.3in
]



\printAffiliationsAndNotice{} 

\begin{abstract}
We propose moment-based variational inference as a flexible framework for approximate smoothing of latent Markov jump processes. The main ingredient of our approach is to partition the set of all transitions of the latent process into classes. This allows to express the Kullback-Leibler divergence between the approximate and the exact posterior process in terms of a set of moment functions that arise naturally from the chosen partition. To illustrate possible choices of the partition, we consider special classes of jump processes that frequently occur in applications. We then extend the results to parameter inference and demonstrate the method on several examples. 
\end{abstract}

\section{Introduction}

\input{introduction}

\section{Background} \label{sec:background}

\input{background}

\section{Moment-Based Variational Smoothing} \label{sec:variational_smoothing}

\input{smoothing}

\section{Parameter Inference} \label{sec:inference}

\input{inference}

\section{Examples}  \label{sec:examples}

\input{examples}

\section{Discussion}

\input{discussion}

\section*{Acknowledgements}

The authors thank the anonymous reviewers for their useful comments and suggestions. This work was supported by the European Research Council (ERC) within the CONSYN project, grant agreement number 773196, and by the Hessian research priority programme LOEWE within the project CompuGene. 

\bibliography{library}
\bibliographystyle{icml2019}

\onecolumn

\appendix

\renewcommand{\theequation}{\thesection.\arabic{equation}}
\setcounter{equation}{0}

\input{supplement}

\end{document}

%% file: introduction.tex
Markov jump processes (MJPs) are popular modelling tools in a number of domains including physics, biology, mathematical finance and chemistry \cite{anderson_1991}. In many cases, these models are used for planning and prediction. To achieve this, the model parameters have to be calibrated from sparse and noisy measurements. Mathematically, this leads to the problem of inference for stochastic processes. Unfortunately, inference in such scenarios is notoriously difficult because parameter inference requires estimation of the latent Markov chain given the observations as an intermediate step. This is especially true in scenarios with unbounded state-spaces such as population models. In recent years, population models have become popular in systems and synthetic biology to describe the intrinsic stochasticity of bio-chemical reactions in a cellular environment \cite{anderson_2015}.  \\
 Algorithms for latent state estimation are often based on the sequential Monte Carlo framework \cite{doucet_2011} or specialized  MCMC approaches \cite{rao_2013}. Alternatively, if the state space is not too large, one can also apply a continuous version of the classical forward-backward algorithm for hidden Markov models.  In the context of parameter learning, many approaches are likelihood-based and apply a form of the EM-algorithm \cite{bladt_2005,metzner_2007,liu_2015}. Bayesian approaches to joint estimation of latent states and parameters usually focus on the particle Markov chain Monte Carlo framework which has inspired many applications \cite{andrieu_2010,golightly_2011,frigola_2013}. \\
While variational inference is ubiquituos in machine learning, most applications focus on probabilistic graphical models \cite{blei_2017}.  An important contribution in the domain of continuous time dynamic models is the work of \citet{archambeau_2007} who developed a Gaussian approximation for stochastic differential equations. \citet{opper_2008} proposed the first variational method for MJPs on the process level. The core idea of their approach is to approximate a coupled multi-component process by a product of independent component processes.  This idea was adapted by \citet{cohn_2010} to continuous time Bayesian networks. More recently, \citet{zhang_2017} combined uniformization with the classical variational method for graphical models. \\ In this work, we propose a general procedure to derive variational inference algorithms for arbitrary MJPs on a countable state space. While based on the same divergence functional as the approach in \cite{opper_2008}, our approximation is conceptually different. The key step in our approach is to partition the state of all transitions into a number of predefined classes. We then express the divergence function in terms of natural moment functions that are induced by the partition. This allows to approximate the inference problem by an optimal control problem with a complexity controlled by the chosen partition for the transitions. Another difference to the mean-field  approach is that we construct our variational family as a modification of the prior process, thus overcoming issues with absolute continuity that can arise from a product approximation. \\ The remaining part of the paper is organized as follows. Section \ref{sec:background} summarizes background material on MJPs. In Section \ref{sec:variational_smoothing} we construct our variational smoothing algorithm and discuss a few special process classes. Section \ref{sec:inference} discusses an extension to latent parameter inference. Finally, in Section \ref{sec:examples} we demonstrate the methods on several examples.

%% file: background.tex
\subsection{ Markov Jump Processes} \label{sec:multi_component_ctmc}

We consider a Markov jump process $ X(t)$ on a countable state space $\mathcal X$ over a finite interval $[0,T]$. As is well-known, such a process is fully specified by an initial measure $p_0: \mathcal X \rightarrow [0,1]$ and a transition function $Q: \mathcal X \times \mathcal X \times [0,T] \rightarrow [0,\infty)$. The transition function defines the infinitesimal transition probabilities in the sense that for $y \neq x$ we have
\begin{equation*}
P( X(t+ \Delta t) = y  \mid X(t) = x) = Q(x,y,t) \Delta t + o( \Delta t) 
\end{equation*}
 For any $t>0$ the marginal probability distribution $p ( x, t) = P( X(t) = x )$ satisfies the master equation
\begin{equation*}
\frac{d}{dt} p(x,t) = \sum_{y \neq x} Q(y,x,t) p(y,t) - \sum_{y \neq x} Q(x,y,t) p(x,t) \, . 
\end{equation*}
Let $g:\mathcal X \rightarrow \mathbb R$ be a function with finite expectation $\mathsf E[ g(X(t) ] < \infty $ for $t \in [0,T]$. Then the expectation obeys a differential equation of the form
\begin{equation}  \label{eq:mjp_moment_dynamics}
\frac{d}{dt}  \mathsf E[ g(X(t) ] = \sum_{y} \mathsf E \left[ \left( g(y) - g(X(t) \right) Q( X(t),y ,t) \right] \, ,
\end{equation}
which can usually not be expressed in terms of $\mathsf E [ g(X(t) ]$ alone \cite{ethier_2005}.  

\subsection{Path Divergence}

For two probability measures $\mu$ and $\nu$  on a common probability space, the Kullback-Leibler (KL) divergence from $\mu$ to $\nu$ is defined as
\begin{equation*}
 \KL{ \mu }{ \nu } = \begin{cases}
  \int \log \left( \frac{d \mu }{ d \nu} \right)  \, d \mu \,  & \mu \ll \nu, \\
  \infty	& \text{otherwise} \, ,
  \end{cases}
\end{equation*}
where $\frac{d\mu}{d \nu }$ is the Radon-Nikodym derivative of measures and $\mu \ll \nu$ refers to $\mu$ being absolutely continuous with respect to $\nu$. \\
Consider two MJPs $X$ and $Z$ on $\mathcal X$ over $[0,T]$ with possibly time-dependent transition functions  $Q^X, Q^Z$ that share the same initial distribution. Let  $P^X $, $P^Z$ denote the measures over the sample path space induced by $X$ and $Z$.  Then the divergence on the level of sample paths is given by
\begin{equation}  \label{eq:mjp_kl}
\begin{split}
 & \KL{P^X  }{P^Z } =  \int_0^T \sum_{x}  p^X (x,t) \sum_{y\neq x} \left[ Q^Z(x,y,t)  \right. \\
 &\quad \left. - Q^X (x,y,t) - Q^X (x,y,t) \log \left( \frac{ Q^X(x,y,t) }{ Q^Z ( x, y,t ) } \right) \right] dt   \, ,
\end{split}
\end{equation}
where and $p^X(\cdot,t)$ refers to the marginal distribution of $X(t)$. For this expression to be finite it is required that $Q^X(x,y,t)$ is zero whenever $Q^Z(x,y,t)$ is zero. A rigorous proof of \eqref{eq:mjp_kl} can be obtained using Girsanov's theorem for counting processes \cite{kipnis_1998}.  Alternative derivations are given in \cite{cohn_2010} based on explicit integration over waiting times or in \cite{opper_2008} using a limit of discrete-time Markov chains for vanishing time steps.

\subsection{Conditional Processes} \label{sec:conditional_processes}

From here on, we focus on a process $X$ with time-independent transition function $Q$. Often the true state  $X(t)$ of the system is only accessible through a $p$-dimensional observation process $Y.$ We focus on the common scenario where observations occur at fixed discrete times $0 \leq t_1, \ldots, t_n \leq T$ that are not necessarily equidistant. Such an observation process may be written as a single matrix-valued random variable  $Y = (Y(t_1), \ldots, Y(t_n))$. We assume that the observation at time $t_k$ only depends on the latent state at time $t_k$ and denote the conditional density of the observations as $p(y_k \mid X(t_k) )$. \\
From a Bayesian perspective, we are interested in the conditional probability distribution $P( X(t) = x \mid y )$ for a particular realization $y \in \mathbb R^{p\times n}$. 
For this type of observation, it can be shown that the smoothed process $\tilde X$ associated with  $P( X(t) = x \mid y )$  is a non-homogeneous MJP \cite{huang_2016}. The corresponding transition function $\tilde Q$ is now time dependent and given by 
\begin{equation} \label{eq:posterior_intensity}
\tilde Q(x,y,t) = \frac{\sigma(y,t)}{ \sigma(x,t) } Q(x,y) \, .
\end{equation}
Here $Q$ is the transition function of the prior process $X$ and $\sigma(x,t) = p(  y_k, \ldots, y_n \mid X(t) = x) $ with $k = \min_{k\in \mathbb N} \{ t_k > t \}$. It can be shown that $\sigma$ obeys the backward equation
\begin{equation} \label{eq:backward_equation}
\frac{d}{dt} \sigma(x,t) = - \sum_{ y \neq x}  Q(x,y) \left( \sigma(y,t) - \sigma(x,t) \right)
\end{equation}
that has to be solved for the terminal condition $\sigma(x,T) = 1$ for all $x\in \mathcal X$ and jump conditions 
\begin{equation*}
\lim_{ t \nearrow t_k}  \sigma(x,t) = \sigma(x,t_k) p(y_k \mid x) \, 
\end{equation*}
for all $x \in \mathcal X$ at the observation times $t_k$. 

\subsection{Variational Smoothing}

Since the smoothed process $\tilde X$ as defined in Subsection \ref{sec:conditional_processes} is typically intractable, we aim to find the best process level approximation $Z$ within a simpler variational class $\mathcal U$. Following the usual variational inference framework \cite{jordan_1999,blei_2017}, we formalize this problem by choosing $Z^* \in \mathcal U$ such that
\begin{equation} \label{eq:variational_problem}
Z^* = \arg \min_{Z \in \mathcal U}  \KL{P^{Z}  }{P^{\tilde X} } \, 
\end{equation}
minimizes the path divergence to the true posterior process. The functional optimization problem \eqref{eq:variational_problem} still depends on the true posterior process $\tilde X$. As shown in the supplement, we can rewrite the objective function in \eqref{eq:variational_problem} as 
\begin{equation} \label{eq:variational_problem_simplified}
\begin{split}
& \KL{P^{Z} }{P^{\tilde X} } = \KL{P^{Z} }{P^{ X} }  \\
 &\quad - \sum_{k=1}^n \mathrm E [ \log p( y_k \mid Z(t_k) ] + \mathrm{const} \, 
 \end{split}
\end{equation}
which is independent of the exact functional form of the posterior intensities. 

%% file: smoothing.tex
\subsection{Variational Process Family} \label{sec:variational_process_family}

In our scenario, the smoothed process $\tilde X$ is an MJP with the modified intensities \eqref{eq:posterior_intensity}. Intuitively, the modified transition rate accounts for deviations of the posterior from the prior process. If the observations suggest that transition $x\rightarrow y$ has occurred more often than expected up to a certain time point, the posterior process will scale up the intensity to match the observation. The core idea of our approximation is to mimic this behavior with a process that is simpler than the full posterior. Thus, we define the variational transition function $Q^Z$ as
\begin{equation} \label{eq:variational_transition_function}
Q^Z(x,y,t) = \lambda(x,y,t) Q(x,y) \, ,
\end{equation}
where we have introduced the variational scaling factor $\lambda$. Note that \eqref{eq:variational_transition_function} corresponds to an overparametrization of the variational problem. Solving \eqref{eq:variational_problem} with the variational family induced by \eqref{eq:variational_transition_function} recovers the true posterior transition function \eqref{eq:posterior_intensity}. In order to achieve a reduction in complexity, we group transitions together such that all transition within the group share a single scaling factor.  \\
To make this more precise, let $\Psi = \{ (x,y)  \in \mathcal X \times X: x \neq y, Q(x,y) > 0 \}$ be the set of all possible transitions of the prior process. Consider a partition $\Pi = \{ \Pi_i \subset \Psi:  i = 1, \ldots, r \} $ such that $\Pi_i \cap \Pi_j = \emptyset $ for $i \neq j$ and $\bigcup_{i=1}^r \Pi_i = \Psi$.  The partition $\Pi$ induces a variational family by restricting the variational scaling factor $\lambda$ to be of the form
\begin{equation} \label{eq:lambda}
\lambda(x,y) = \lambda_i (t)  \quad \text{for} \quad  (x,y) \in \Pi_i \, ,
\end{equation} 
where the $\lambda_i:[0,T] \rightarrow [0,\infty)$ are from a suitably regular class of functions. Systematic ways to choose such a transition space partition will be discussed in Section \ref{sec:special_cases} and Section \ref{sec:inference}. 

\subsection{Moment-Based Divergence } \label{sec:moment_based_divergence}

Define the natural moment functions associated with the partition $\Pi$ as
\begin{equation} \label{eq:natural_moment}
\varphi_i(t) =  \sum_{ (x,y) \in \Pi_i   } Q(x,y) p^Z(x,t) \, ,
\end{equation}
where $p^Z$ refers to the  marginal distribution of the variational process $Z(t)$. It is also convenient to introduce the function $\tilde Q_i$ as
\begin{equation*}
\tilde Q_i (x) = \sum_{y} \mathbf 1_{\Pi_i} (x,y) Q(x,y) \, 
\end{equation*}
with the indicator function $\mathbf 1_{\Pi_i} (x,y)=1 $ iff the transition $(x,y)$ is contained in $\Pi_i$. Intuitively, $\tilde Q_i(x)$ corresponds to the total exit rate from $x$ to states within transition class $i$. We can now express the natural moment functions as
\begin{equation*} 
\varphi_i(t) = \mathsf{E} [ \tilde Q_i( Z(t) ) ]  \, .
\end{equation*}
Consider the divergence from the variational process to the prior process $D[ P^Z \, || \, P^X ] $ in \eqref{eq:variational_problem_simplified}. Exploiting \eqref{eq:variational_transition_function}, \eqref{eq:lambda} and \eqref{eq:natural_moment} we obtain
\begin{equation*}
D[ P^Z \, || \, P^X ] = L[ \varphi, \lambda]  
\end{equation*}
where $\varphi$ and $\lambda$ are vector-valued functions with components $\lambda_i$ and $\varphi_i$, respectively.  The functional $L$ is defined as
\begin{equation*}
 L[ \varphi, \lambda]  = \sum_{i=1}^r \int_0^T \varphi_i(t) \left( 1 - \lambda_i (t) + \lambda_i(t) \log \lambda_i(t)  \right) dt \, .
\end{equation*}
Since the variational process $Z$ is a non-homogeneous MJP, the moment functions $\varphi$ and the scaling factors $\lambda$ are not independent, but obey a differential equation of the form \eqref{eq:mjp_moment_dynamics}
\begin{equation}  \label{eq:natural_moment_dynamics}
\begin{split}
 &\frac{d}{dt} \varphi_i(t)  = \sum_{j=1}^r \lambda_j(t) \mathsf{E}  \left[ \sum_{y} \left( \tilde Q_i(y) \right. \right. \\
 &\quad \left. \left. - \tilde Q_i(Z(t)) \right) \mathbf 1_{\Pi_j} ( Z(t),y) Q(Z(t) ,y)   \right] \, .
 \end{split}
\end{equation}
Our goal is to express the r.h.s. of \eqref{eq:natural_moment_dynamics} in terms of $\varphi(t)$, $\lambda(t)$ to obtain a closed-form description of the variational problem in terms these quantities. In general, this is not possible and we obtain an equation of the form
\begin{equation} \label{eq:natural_moment_dynamics_reduced}
\frac{d}{dt} \varphi (t)  = A( \lambda (t) ) \varphi (t) + B (\lambda (t) ) \tilde \varphi (t)
\end{equation}
where $A$ and $B$ are suitable matrix valued functions and $\tilde \varphi $ is a collection of higher order moment functions that cannot be reduced to $\varphi$. In this case, we can apply an additional approximation in the form of moment closure as often used in the analysis of non-linear stochastic dynamical systems \cite{kuehn_2016}. Generally speaking, a moment closure scheme is a function $\psi$ such that we may write $ \tilde \varphi(t) = \psi ( \varphi (t) ) $. Thus, we arrive at an equation of the form
\begin{equation} \label{eq:forward_equation}
\frac{d}{dt} \varphi (t)  = f( \lambda(t), \varphi (t) ) \, ,
\end{equation}
where $f$ depends on the process and the applied closure scheme. We can now recast the variational inference problem \eqref{eq:variational_problem} into a non-linear optimal control problem of the form
\begin{equation}  \label{eq:control_problem}
\begin{aligned}
\mathrm{minimize} & &  &L[ \lambda, \varphi] - F[\varphi] \\
\mathrm{subject \ to} & &  &\frac{d}{dt} \varphi(t) = f( \lambda(t), \varphi(t) )    \, . \\   
\end{aligned} \, 
\end{equation}
Here, $F$ refers to the observation contribution in \eqref{eq:variational_problem_simplified}, i.e. 
\begin{equation*}
F[ \varphi ] = \sum_{k=1}^n \mathsf E[ \log p(y_k \mid Z(t_k) ] 
\end{equation*}
which has to be understood as a functional of the natural moments $\varphi$.  Before turning to special classes of MJPs, we end this section with a remark on moment closure. When the moment equations \eqref{eq:natural_moment_dynamics_reduced} are naturally closed (i.e. $B=0$), than the control problem \eqref{eq:control_problem} corresponds exactly to the process level formulation \eqref{eq:variational_problem} and the minimum of $L$ corresponds to the usual evidence lower bound. When we use moment closure, there is no global process equivalent to \eqref{eq:control_problem} and the lower bound property is lost. This effect has also been observed for other variational approximations that go beyond product mean-field, for example cluster variational methods \cite{yedidia_2000}. In our case, we can recover a valid process level result using the solution $\lambda^*$ in \eqref{eq:variational_transition_function}. Generating sample paths from the so defined approximate posterior gives a means to analyze the accuracy of the moment closure approximation. 

\subsection{Special Classes} \label{sec:special_cases}

In this section, we consider special classes of structured MJPs that give rise to  a natural partition of the transition space. 

\paragraph{State Space Lumping.}

Consider an MJP $X$ with countable state space $\mathcal X $ and let $ \{ \Pi_1, \ldots, \Pi_s \}$ be a partition of the state space (in contrast to the partition of the transition space). The lumped process can now be defined as
\begin{equation*}
 S (t) = j \quad  , \quad \text{for }  X(t)=x , \  x \in \Pi_j  \, .
\end{equation*}
A partition of the state space naturally induces a partition of the transition space given by
 \begin{equation*}
 \Pi_{ij} = \{ (x,y) \in \{1, \ldots, s \}^2 : x \in \Pi_i , y \in \Pi_j \} \, .
 \end{equation*}
 This partition subsumes all transitions of the lumped process. Accordingly, the natural moment functions become
 \begin{equation*}
 \varphi_{ij}(t) = \sum_{x,y} \mathbf 1_{\Pi_i} (x) \mathbf 1_{\Pi_j} (y) Q(x,y) q(x,t) \, .
 \end{equation*}
It is also interesting to observe that
\begin{equation*}
\tilde Q_{ij} (x) =  \mathbf 1_{\Pi_i} (x) \sum_y \mathbf 1_{\Pi_j} (y) Q(x,y) 
\end{equation*}
which is the total transition rate from $x$ in $\Pi_i$ to $\Pi_j$. In the special case that the MJP is lumpable with respect to the partition $\Pi$ \cite{rubino_1993}, we obtain 
\begin{equation*} 
\tilde Q_{ij}(x)  = \begin{cases}
\tilde Q_{ij}  & x \in \Pi_i \\
0 & \text{otherwise} 
\end{cases} \, ,
\end{equation*}
where $\tilde Q_{ij} = \sum_{y \in \Pi_j} Q(x,y)$  which is identical for all $x \in \Pi_i$. 
Inserting this into the dynamic equation yields
\begin{equation} \label{eq:lumping_master_equation}
\frac{d}{dt} \varphi_{ij} = \tilde Q_{ij}  \left[ \sum_{l \neq i} \lambda_{li} (t) \varphi_{il} (t) - \sum_{l \neq i} \lambda_{il} (t) \varphi_{il} (t)  \right]
\end{equation}
Dividing both sides by $\tilde Q_{ij}$, we observe that \eqref{eq:lumping_master_equation} is equivalent to the master equation of the lumped process $S$. Accordingly, our variational approach will recover the exact posterior of the lumped process. While most examples of practical interest are not lumpable, \eqref{eq:lumping_master_equation} provides a way to exploit known approximate lumping schemes for inference purposes by understanding the approximate lumping as a moment closure scheme. 

\paragraph{Data Driven Partitioning.}

In many cases, the observations $Y$ do not depend on the full latent state $X(t)$ but rather on a summary statistic $S(t) = T(X(t))$ for $T: \mathcal X \rightarrow \mathcal S$.  In general, $S(t)$ will be a non-Markovian jump process on $\mathcal S$. However, it induces a natural partition of the form
\begin{equation*}
\Pi_{z,z'} = \{ (x,x') \in \Psi : T( x ) = z, T(x') = z' \} 
\end{equation*}
meaning that we group all transitions of $X$ that lead to the same state change of $S$ in $\mathcal S$. 

 \paragraph{Agent-Based MJPs.}
 
 An agent-based MJP $X$ on $\mathcal X = \mathbb \{ 1, \ldots, m \}^d$ describes the behavior of $d$ coupled agents where each agent may be in one of $m$ states. In addition, the probability of two agents changing the state at the same time is assumed to be zero.  In terms of the transition function, we obtain
 \begin{equation*}
 Q(x,y) = \begin{cases}
  Q^i_{jk} (x)  &  x_i = j, y_i = k, x_l = y_l \text{ for } l \neq  i \\
  0 & \text{otherwise} 
  \end{cases}
 \end{equation*}
 where $Q^i_{jk}$ is the transition rate of agent $i$ from state $j$ to state $k$ as a function of the state of the other agents. This is the class of processes targeted by the existing mean-field approaches for jump processes \cite{opper_2008,cohn_2010}. Since our variational family is constructed by modifying the prior process, we cannot recover the product-type approximation used in the classical approaches. We can, however, mimic its behavior by subsuming all transitions that produce the same change of the same agent, i.e.  
\begin{equation} \label{eq:agent_parttion}
\Pi^i_{jk} = \{ (x,y) \in \Psi : x_i = j, y_i = k, x_l = y_l  \} \, .
\end{equation}
Using \eqref{eq:agent_parttion}, we can evaluate the corresponding moment functions as
\begin{equation*} 
\varphi^i_{jk} (t)  = \mathsf E [ \mathbf 1_{j} ( Z_i(t) ) Q^i_{jk} ( Z(t) ) ] \, ,
\end{equation*}
which is quite intuitive, since it is the transitions rate of agent $i$ to go from $j$ to $k$ averaged over all configurations of the remaining agents. 
 
 \paragraph{Population-Based MJPs.}
 
 A population-based MJP $X$ on $\mathcal X = \mathbb N_0^d$ describes the stochastic evolution of the abundances of $d$ species over time.  While they seem similar to agent models from the previous section, population models usually have an unbounded state space. In addition, one event may affect several species at the same time. This fact  is formalized by a set of change vectors $\{ v_i  \in \mathbb Z_0^d: i = 1, \ldots, r \}$ and a transition function of the form
 \begin{equation*}
 Q(x,y) = \begin{cases}
 h_i (x) & \text{ if } y = x+v_i \\
 0 & \text{ otherwise}
 \end{cases} \, .
 \end{equation*}
If there are change vectors that affect several species, the product mean-field approach is ill-suited for this type of system, since a product of independent processes will not be absolutely continuous with respect to the prior process.  In our framework, a natural choice of partition for this type of model is to combine all transitions that correspond to the same change vector
 \begin{equation*}
 \Pi_i = \{ (x,y) \in  \Psi : y = x+v_i \} \, .
 \end{equation*}
 The corresponding natural moment functions become
\begin{align*}
\varphi_i(t) &= \mathsf E [ h_i (Z (t)) ] \, .
\end{align*}
The time-evolution of $\varphi $ is given by
\begin{equation*}
\begin{split}
\frac{d}{dt} \varphi_i (t) &= \sum_{j} \lambda_j(t) \mathsf E \left[  \left(  h_i( Z(t) + v_j ) \right. \right. \\
&\quad \left. \left. - h_i ( Z(t) )  \right) h_j (Z(t)  \right] \, .
\end{split}
\end{equation*}
For systems where $h_i$ are linear functions of the state, this equation is closed in $\varphi_i$. An example of this class will be discussed later. 

\subsection{Minimizing the Divergence.}

We now introduce an algorithm to find approximately optimal variational scaling functions $\lambda^*$. We do this following the indirect approach from optimal control \cite{bonnard_2006}. The idea here is to transform the constrained problem \eqref{eq:control_problem} into an unconstraint problem by introducing the Lagrangian
\begin{equation*} \label{eq:control_problem_unconstrained}
\begin{split}
J [ \lambda, \varphi, \eta] &= L[\lambda,\varphi] - F[\varphi]  \\
&\quad -\int_0^T \eta(t)^T \left[ f( \lambda (t) , \varphi(t) ) - \dot \varphi (t)  \right] \, ,
\end{split}
\end{equation*}
where $\eta$ is a vector-valued Lagrange multiplier function. By taking the functional derivative with respect to $\varphi_i$ we obtain 
\begin{equation} \label{eq:costate_equation}
\begin{split}
\frac{d}{dt} \eta_i (t) &=  \sum_{j=1}^r \left( 1 - \lambda_j(t) + \lambda_j(t) \log \lambda_j (t) \right) \\
&\quad-  \sum_{j=1}^r  \frac{d f_j }{ d \varphi_i} \eta_j(t) \, 
\end{split}
\end{equation}
valid in between the observations. At the point of the observations, the functional $F$ will induce jump conditions for $\eta$ given by
\begin{equation} \label{eq:costate_reset}
 \lim_{t \nearrow t_k }\eta_i(t) = \eta_i(t_k) + \frac{d}{ d \varphi_i(t_k) } \mathsf E[ p(y_k \mid Z(t_k ) )  ]  \, .
\end{equation}
Similarly, differentiation with respect to the scaling factor $\lambda_i$ yields an algebraic constraint of the form
\begin{equation} \label{eq:control_condition}
0 = \varphi_i(t) \log \lambda_i (t) - \sum_{j=1}^r \eta_j(t) \frac{ d f_j }{ d \lambda_i } \, .
\end{equation}
In the classical mean-field algorithm, the scaling functions $\lambda$ can be expressed in terms of $\eta$ by solving \eqref{eq:control_condition}. The control problem is then solved using a form of the forward-backward sweep method \cite{mcasey_2012}. In our framework, it is generally  not possible to eliminate the scaling factors. A variation of the forward-backward sweep explicitly including the scaling factors turned out to be unstable in our experiments.  \\
We therefore propose a gradient-based algorithm based on the following argument. The moment functions $\varphi$ are fully determined by the scaling factors $\lambda$ and an initial condition. Hence, we may understand $L$ as a functional of $\varphi$ only. In this case, the r.h.s. of \eqref{eq:control_condition} corresponds to the gradient of $L$ with respect to $\varphi$. Since the raw gradient updates obtained this way do not work too well, we take advantage of the probabilistic nature of the objective function to derive an analogue to the natural gradient \cite{amari_1998}. We do this by performing an expansion of the path divergence up to second order in $\lambda$. From this, we obtain updates of the form
\begin{equation} \label{eq:gradient_step}
\lambda^{(n+1)} =\lambda^{(n)} -h \tilde \nabla L[ \lambda^{(n)} ]  
\end{equation} 
 with step size $h$ and the and the natural gradient
 \begin{equation} \label{eq:natural_gradient}
 \tilde \nabla L[\lambda] =   \lambda_i(t)  \log \lambda_i (t)   - \frac{\lambda_i(t) }{ \varphi_i (t) } \sum_{j=1}^r \eta_j(t) \frac{ d f_j }{ d \lambda_i } \, .
\end{equation}  
 The updates \eqref{eq:gradient_step}, \eqref{eq:natural_gradient} ensure that we take small steps on the manifold defined by the variational family and improve convergence significantly. The resulting algorithm is summarized in Algorithm \ref{alg:grad_mbvi}.  A more detailed discussion including a derivation of the natural gradient can be found in the supplement Section 3.

  \begin{algorithm}[t]
   \caption{Natural Gradient Descent for MB-VI}
   \label{alg:grad_mbvi}
\begin{algorithmic}[1]
   \STATE {\bfseries Input:} Initial guess for the scaling factors $\lambda^{(0)}$, \\ initial condition $\psi(0)$.
   \REPEAT
   \STATE Given $\lambda^{(n)}$, $\varphi(0)$, compute $\varphi^{(n)}$ using \eqref{eq:forward_equation}.
   \STATE Given $\lambda^{(n)}$, $\varphi^{(n)}$, compute $\eta^{(n)}$ using \eqref{eq:costate_equation}, \eqref{eq:costate_reset}.
   \STATE Given $\lambda^{(n)}$, $\varphi^{(n)}(t)$ and $\eta^{(n)}$, evaluate the natural gradient $ \tilde \nabla L $ according to \eqref{eq:natural_gradient}. 
   \STATE Set $\lambda^{(n+1)}$ according to \eqref{eq:gradient_step}.
   \UNTIL{ $ | L [ \lambda^{(n)} ] - L [ \lambda^{(n-1)} ] | < \mathrm{tolerance}$  }
   \STATE {\bfseries Output:} Optimized variational scaling factor $\lambda^*$ .
\end{algorithmic}
\end{algorithm}

%

%% file: inference.tex
Suppose now that the transitions function $Q$ of the prior process depends on a collection of parameters $\theta$. Using the usual framework of the variational EM algorithm, we can find an approximate maximum likelihood estimate by understanding the objective function $L[\lambda,\theta]$ as a function of the scaling factors and the parameters and then minimizing iteratively with respect to both arguments. In general, not much can be said about the structure of the resulting optimization problem. In the following, we focus on a special case that allows for closed form parameter updates. \\
Let $\Pi$ be a partition of the transition space such that the prior process has parametrized transition function of the form
\begin{equation*}
Q_\theta (x,y ) = c_i( \theta)  h(x,y) \quad \text{for} \quad (x,y) \in \Pi_i \, ,
\end{equation*}
where $c_i$ and $h$ are known functions with $c_i$ differentiable with respect to $\theta$. We proceed by defining our variational family on $\Pi$ by setting
\begin{equation} \label{eq:parameter_variational_family}
\lambda(x,y,t) = \lambda_i(t) h(x,y) \quad \text{for} \quad (x,y) \in \Pi_i \, . 
\end{equation}
Note that \eqref{eq:parameter_variational_family} slightly differs from the ansatz introduced in Section \ref{sec:variational_process_family} in the fact that it only contains a part of the prior intensity function. As a consequence, the variational family remains independent of the parameters $\theta$. Setting 
\begin{equation*}
\varphi_i(t) = \sum_{(x,y) \in \Pi_i}  h(x,y) q(x,t) \, ,
\end{equation*}
we can reuse the formalism introduced in section \ref{sec:moment_based_divergence} to obtain stationarity conditions for $\lambda$. In addition, for fixed $\lambda$, $\varphi$ we obtain optimality conditions for the parameters of the form
\begin{equation*}
\sum_{j}  \left( G_j c_j(\theta) - H_j \right) \frac{d c_j (\theta)  }{d \theta_i }  = 0 \,  ,
\end{equation*}
where we have introduced the summary statistics
\begin{equation*}
\begin{split}
G_i &= \int_0^T \varphi_i(t) dt \, , \\
H_i &= \int_0^T \varphi_i(t) \lambda_i (t) dt \, . 
\end{split}
\end{equation*}
Intuitively, $H_i$ corresponds to the expected number of transitions in class $i$ during $[0,T]$. In the case where we have a single parameter per transition class, i.e. $c_i(\theta)=\theta_i$, we get
\begin{equation*}
\theta_i = \frac{H_j}{ G_j} \, . 
\end{equation*}
As an alternative to the EM approach, we may also follow a Bayesian framework, i.e. by assuming a prior $p(\theta)$. Choosing a variational ansatz that factorizes over parameters and latent trajectories, we get an exponential family variational parameter distribution of the form
\begin{equation} \label{eq:parameter_variational_posterior}
q(\theta) \propto q(\theta) \prod_{j} c_j(\theta)^{H_j} \exp \left( - \sum_{j} c_j ( \theta ) G_j \right) \, .
\end{equation}
If the prior is a gamma distribution and the we have a single parameter per transitions class, \eqref{eq:parameter_variational_posterior} will also have the form of a gamma distribution.

%% file: examples.tex
In this section, we apply our method to three examples. We focus on models of the population type for which inference is notoriously difficult due to the unbounded state space. First we consider a linear birth death process. This simple example allows analytic treatment and provides some intuition. Next, we numerically study a stochastic gene expression model of the type that is frequently used in systems biology. As a third example, we consider a stochastic predator prey model to demonstrate our method in combination with moment closure. 

\subsection{Birth Death Process}

\begin{figure}[h]
\includegraphics[width=\linewidth , trim = 40 220 50 210, clip]{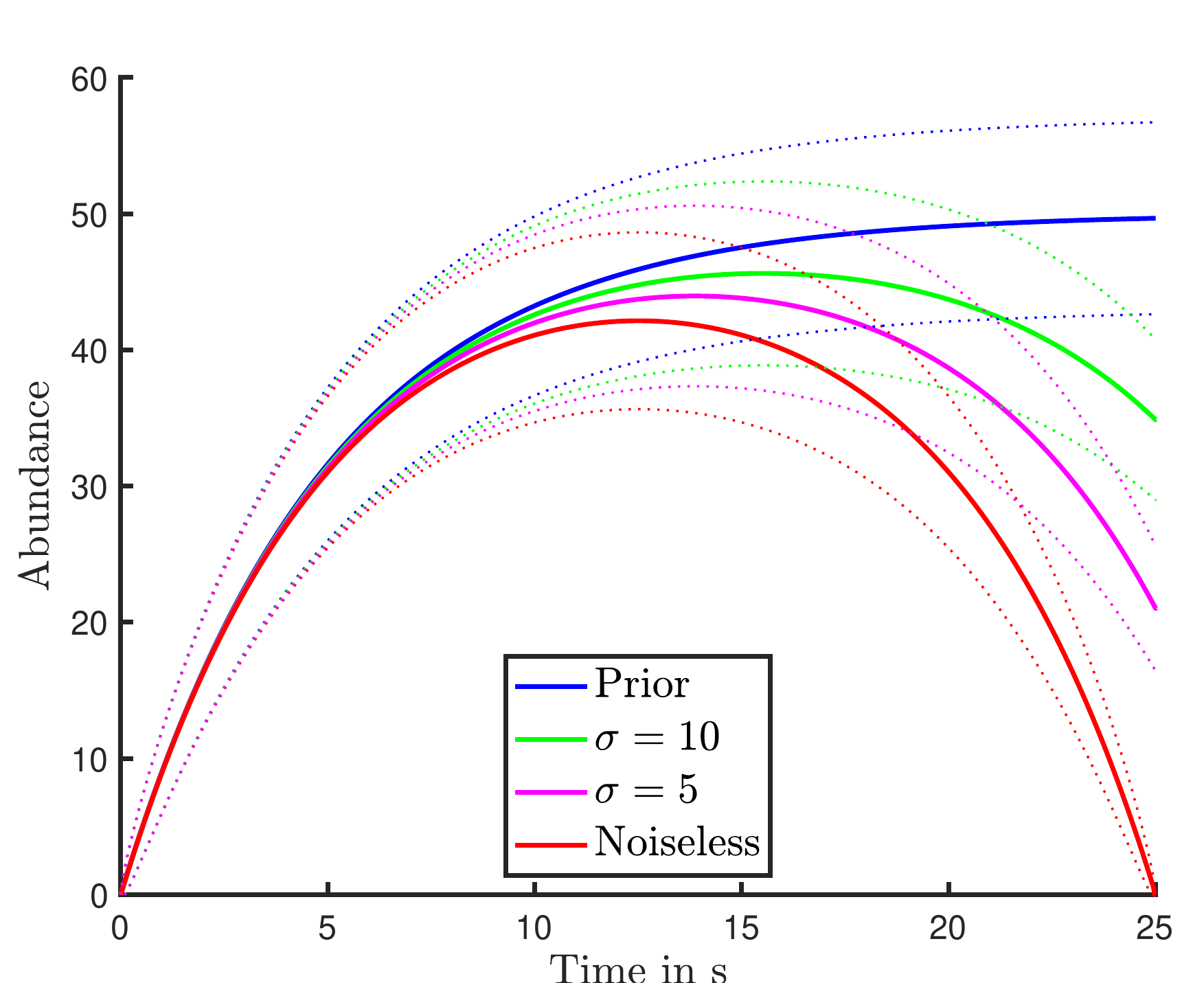}
\caption{ Mean (thick lines) and standard deviation around the mean (dotted lines) for different endpoint constraints. All results used the parameter configuration $c_1 = 5 \mathrm{s}^{-1}$,  $c_2 = 0.1 \mathrm{s}^{-1}$. } 
\label{img:birth_death}
\end{figure}

Let $ \mathcal X = \mathbb N_0$, $X_0\in \mathcal X$ and $c_1,c_2 > 0$. For the transition vectors $v_1 = 1$, $v_2 = -1$ together with the intensity 
\begin{equation*}
Q(x,y) = \begin{cases}
c_1 & y = x+1 \\
c_2 x & y= x-1 \\ 
\end{cases}
\end{equation*}
the resulting process $\{X(t),t\geq 0\}$ is known as the linear birth death process and can be seen as the simplest case of a population-type model. We therefore following the approach of choosing the partition $\Pi = \{\Pi_1, \Pi_2 \}$ with
\begin{align*}
\Pi_1 = \{ (x,y) : y = x+1\}  \ , \ \Pi_1 = \{ (x,y) : y = x-1 \}
\end{align*}
which induce the moment functions
\begin{align*}
\varphi_1(t) = c_1 \ , \ \varphi_2(t) = c_2 \mathsf{E} [Z(t) ] \, . 
\end{align*}
Since, $\varphi_1$ is constant, the moment dynamics are reduced to a single closed equation of the form
\begin{equation} \label{eq:birth_death_moment}
\frac{d}{dt} \varphi_2(t) = c_2 \lambda_1(t) \varphi_1 - c_2 \lambda_2(t) \varphi_2(t) \, .
\end{equation}
Now suppose we consider the conditional process with endpoint condition $X(T) = 0$. For simplicity, let us also assume that $X_0 = 0$.  In this case, the backward equation can be integrated analytically. This allows to compute the variational scaling factors as  
\begin{align*}
\lambda_1(t) &=  (1- \exp( T-t) ) , \\
\lambda_2(t) &=  (1- \exp( T-t) )^{-1} \, . 
\end{align*}
Inserting this into \eqref{eq:birth_death_moment}  we obtain
\begin{equation*}
\mathsf E[ Z(t) ]  = \frac{c_1}{c_2} ( 1 - \exp[-c_2 (T-t) ] ) (1 - \exp (-c_2 t) ) \, .
\end{equation*}
For this example, the true posterior intensities can be calculated analytically \cite{huang_2016} leading to the same result. Hence, the variational approximation coincides with the exact smoothed process in this simple special case. A graph of the first moment and the corresponding standard deviation can be found in Fig. \ref{img:birth_death}.  \\
Now consider the same example where instead of a terminal condition we have obtained a single observation $Y_T = y$ with zero mean additive Gaussian noise of variance $\sigma^2$. In this case, the objective function becomes more complicated, since the second order moment has to be included in order to describe the observations. Consequently, there seems to be no simple closed form expressions to characterize the variational posterior. We ran our algorithm for different values of $\sigma^2$ and verified empirically that the solution converges to the analytic expression for the exact terminal constraint case with $\sigma \rightarrow 0$. Fig. \ref{img:birth_death} shows the mean of the prior process and the noise-free conditional process compared two two observations with different noise levels. As expected, the variational method seems to interpolate between the observation and the prior process. 

\subsection{Gene Expression System}

\begin{figure}[t!]
\includegraphics[width=\linewidth , trim = 40 220 50 210, clip]{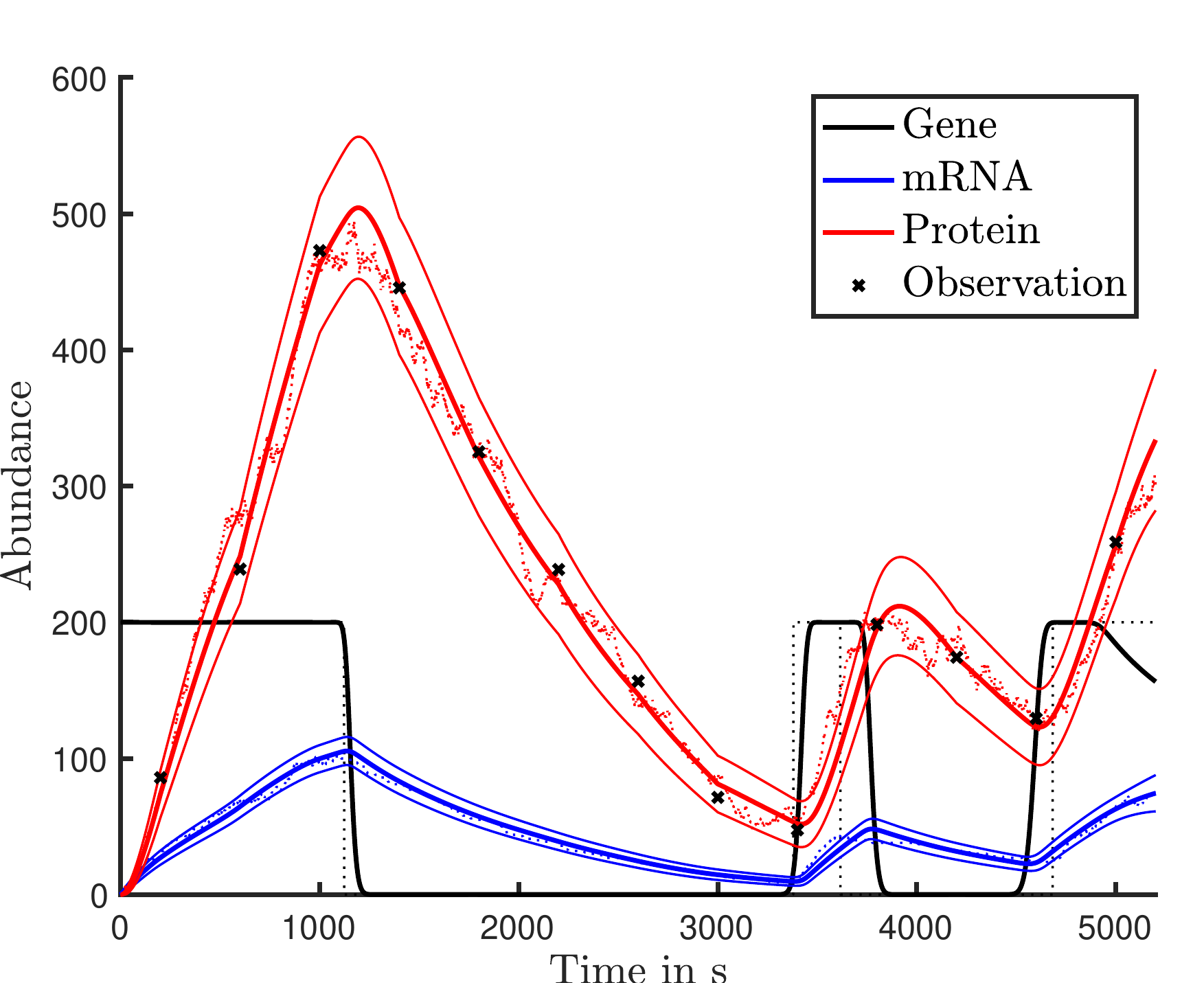}
\caption{Approximate smoothing result for simulated observations from a gene expression model. Thick bold lines denote the mean of the variational posterior process, while the thin lines indicate an interval of one standard deviation around the mean. The true latent trajectories are shown as dotted lines. Note that we only considered a single gene but scaled the curve in the plot for better visibility.}
\label{img:gene_expression_smoothing}
\end{figure}

Consider the following simple gene expression model consisting of a single gene that can switch between an inactive state $G^0$ and an active state $G_1$ with rates $c_1$ and $c_2$. While the gene is active, mRNA is produced at rate $c_3$ (transcription). The mRNA is then processed by ribosomes to produce the target protein at rate $c_5$ (translation). Both mRNA and protein molecules undergo degradation reactions with rates $c_4$ and $c_6$ respectively. \\
The stochastic behavior of this system is modeled by a  population MJP $X$ with state space $\mathcal X \subset \mathbb N_0^3$ where the components correspond to the numbers of active genes $X_1$, mRNA molecules $X_2$ and protein molecules $X_3$. The nonzero elements of the transition function $Q$ are given by
\begin{equation*}
\begin{aligned}
Q(x,x+e_1) &= c_1(x_1-1) , &  Q(x,x-e_1) &= c_2 x_1 , \\
Q(x,x+e_2) &= c_3 x_1 , &   Q(x,x-e_2) &= c_4 x_2 , \\
Q(x,x+e_3) &= c_5 x_2 , &   Q(x,x-e_3) &= c_6 x_3 .
\end{aligned}
\end{equation*}
In typical applications, the produced protein molecule is a fluorescent reporter. Measurements of the reporter concentration can be obtained from live-cell microscopy by integrating fluorescence intensity over the cross-section of the cell. Thus, these measurements are a scaled and noisy observation of the process component corresponding to the protein abundance. For simplicity, we assume that the noisy observations $Y_1, \ldots, Y_n$  are conditionally Gaussian given the protein abundance. To construct our variational approximation, we choose the population-type partitioning and obtain closed form moment equations that can be found in Section 5 of the supplement.  \\
 The results for variational posterior mean and variance are shown in Fig. \ref{img:gene_expression_smoothing}. The example demonstrates that the posterior mean concentrates on the true realization of the latent process. Thus, the method allows to infer latent state dynamics, in particular, the unobserved activation patterns of the gene are recovered quite well. To investigate this further, we used the posterior mean of the gene state to devise a detector (threshold: 0.5) and evaluated the true positive fraction $\alpha = 0.94$ and false positive fraction $\beta = 0.15$ from $N=100$ pooled trajectories. We note that false detections are mainly caused by short events. For example, if the gene is briefly inactive during a longer period of activity, this may not be visible in the protein activity and even less so in the observations.  \\
 Fig \ref{img:gene_expression_smoothing} also reveals a drawback of our method. From an exact smoothing, we would expect the variance of the posterior to be smaller at times of observations than in between the observations. We do not observe this behavior in our approach. The most likely reason for this is that a single time-dependent factor per reaction channel does not provide enough degrees of freedom to capture this effect. In the end, our approach reduces the full smoothing problem to a set of nine ordinary differential equations. 
 

\subsection{Predator-Prey Dynamics}

\begin{figure}[t!]
\includegraphics[width=\linewidth , trim = 40 220 50 210, clip]{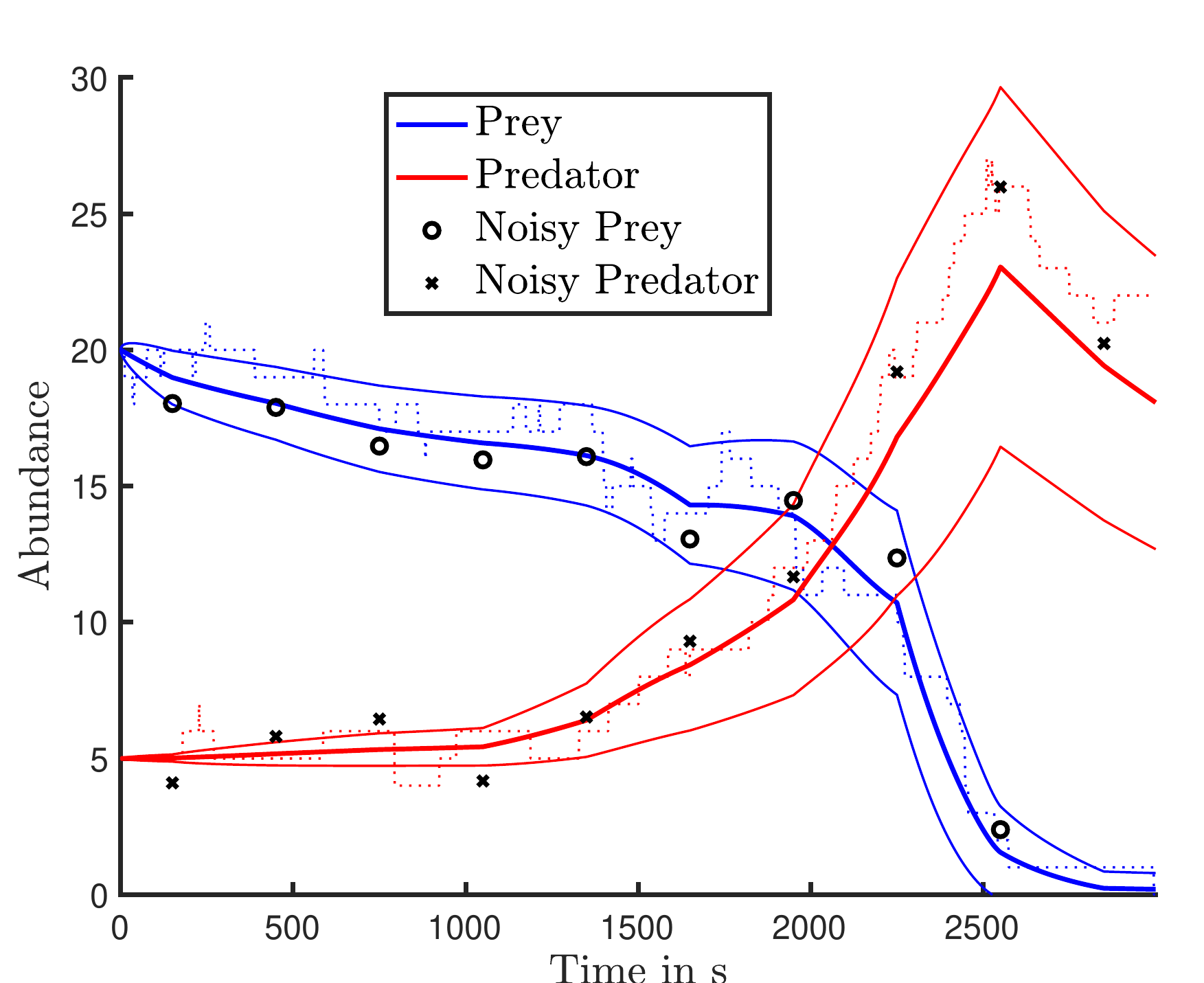}
\caption{ Approximate smoothing result for simulated observations from a stochastic predator prey model. Thick bold lines denote the mean of the variational posterior process, while the thin lines indicate an interval of one standard deviation around the mean. The true latent trajectories are shown as dotted lines.  } 
\label{img:predator_prey}
\end{figure}

To demonstrate our approach on a non-linear model, we consider a stochastic form of the classical predator prey interaction model. Here, a prey species $X_1$ and a predator species $X_2$ interact with each other as defined by the transition function $Q$ with non-zero elements
\begin{equation*}
\begin{aligned}
Q(x,x+e_1) & = c_1 x_1 , &   Q(x, x-e_1) &= c_2 x_1 x_2 , \\
 Q(x,x+e_1) & = c_3 x_1 x_2 , &  Q(x, x-e_1) &= c_4 x_2  . 
\end{aligned}
\end{equation*}
Note that this system is unstable in the sense that, in the long run, we observe either extinction of both species or extinction of the predator and explosion of the prey population. Using our population-type partitioning, we obtain a set of moment equations that is not closed. In Supplement Sec. 5, we demonstrate how moment closure can be employed to obtain a closed system. \\
An application of the variational smoother to synthetic data of the predator prey model is given in Fig. \ref{img:predator_prey}. Due to the low abundances, it is possible to perform a truncation of the state space and solve the smoothing problem by integrating \eqref{eq:backward_equation} backward in time. Comparing Fig. \ref{img:predator_prey} to exact smoothing results reveals that the variance of the approximate posterior is higher than the variance of the exact smoothed process.  

%
%
%

%% file: discussion.tex
On the basis of a partitioning of the transition space, we proposed a flexible framework to construct variational inference procedures for Markov  jump processes. By defining the variational family as a modification of the prior process, our approach circumvents problems with absolute continuity that have plagued earlier product-type mean-field approaches. The main advantage of our method is, however, that the complexity of the approximation depends on the chosen partition of the transition space and can thus be adapted to the problem at hand. To get some insight into the choice of partition, we discussed classes of structured MJPs that give rise to a natural partitioning.  As other variational procedures, moment based variational inference can be used for approximate parameter inference. An interesting observation here is that a suitable choice of partitioning may naturally lead to a gamma variational distribution of the parameters.  Finally, we demonstrated the method on synthetic examples of the population type. \\
Our work opens several lines of future research. First of all, we envision applications to real data, e.g. in the context of systems and synthetic biology. On the theoretical side, it may be interesting to assess the effect of different partition and  moment closure choices on the approximation quality. Finally, a more systematic comparison with related methods is required. \\

%% file: supplement.tex
\section{Additional Derivations}

\subsection{Decomposition of the Divergence}

Here, we show that the divergence between an approximate process and a posterior process can be decomposed into the divergence of the approximate process and the prior process plus a contribution of of the observations. More specifically, we prove \eqref{eq:variational_problem_simplified}.

\begin{lemma} \label{thm:posterior_divergence}
The divergence in \eqref{eq:variational_problem_simplified} can be written as
\begin{equation*}
 \KL{P^{Z} }{P^{\bar X} } = \KL{P^{Z} }{P^{ X}}  - \sum_{k=1}^n \mathsf E [ \log p( y_k \mid Z(t_k) ] + \mathrm{const}
\end{equation*}
and hence the variational problem is independent of the exact functional form of the posterior intensities. 
\end{lemma} 
To see this, let $Z(t)$ be a MJP with marginal distribution $p^Z$ and time-dependent transition function $\lambda$. To simplify the derivation, it is useful to introduce operators $\mathcal L, \mathcal L^\dagger$ defined by their action on a function $f$ of a suitable class
\begin{equation} \label{eq:generator}
 \begin{split}
\mathcal L f(x) &= \sum_{y\neq x} \lambda(y,x,t) f(y) - \sum_{y\neq x} \lambda(x,y,t) f(x) \, , \\
\mathcal L^\dagger f(x) &= \sum_{y\neq x}  (f(y) -f(x)) \lambda(x,y,t) \, .
\end{split}
\end{equation} 
 In analogy to\eqref{eq:mjp_moment_dynamics} we get for a function $g:\mathcal X \times [0,T] \rightarrow \mathbb R$,
\begin{equation} \label{eq:time_dep_backward_equation}
\frac{d}{dt} \mathsf{E} [ g( Z(t), t) ] = \mathsf{E} [ \mathcal L^\dagger g(Z(t), t) ] + \mathsf E[ \partial_t g(Z(t), t ) ] 
\end{equation}
where the second expression comes from the explicit time dependence of the function $g$.  Note that \eqref{eq:time_dep_backward_equation} can be obtained by inserting the distribution function $p^Z$, using the product rule of differentiation and inserting the master equation obeyed by $p^Z$. Following the notation of the main text, we will denote the time independent transition function of the prior process by $Q$. 
Starting from the divergence of two time dependent MJPs \eqref{eq:mjp_kl} and inserting the posterior intensities yields
\begin{equation} \label{eq:kl_decomposition}
\KL{P^{Z}}{P^{\bar X}}  = J_1- J_2 + J_3 + J_4
\end{equation}
with 
\begin{equation*}
\begin{split}
J_1 &= \sum_{y} \int_0^T \mathsf E \left[  \frac{\sigma ( y,t) }{ \sigma ( Z (t), t) }  Q( Z(t) , y)  \right]  dt \, , \\
J_2 &= \sum_{y} \int_0^T \mathsf E \left[  \lambda( Z(t) , y,t) \right] dt  \, \\
J_3 &= \sum_{y} \int_0^T \mathsf E \left[  \lambda( Z(t) , y,t)  \log \left(  \lambda( Z(t) , y,t)   \right) \right] dt \, \\
J_4 &= \sum_{y} \int_0^T \mathsf E \left[  \lambda( Z(t) , y,t)  \log \left( \frac{  \sigma ( Z(t), t ) }{  \sigma ( y, t ) }  \right) \right]  dt
\end{split}
\end{equation*}
Now observe that by \eqref{eq:generator} the expectation within $J_4$ can be written as
\begin{equation} \label{eq:J4}
\sum_{y} \mathsf E \left[  \lambda( Z(t) , y,t)  \log \left( \frac{  \sigma ( Z(t), t ) }{  \sigma ( y, t ) }  \right) \right]   =   - \mathsf E [ \mathcal L^\dagger \log \sigma(Z(t),t) ]  \, .
\end{equation}
By addition of a zero, we can rewrite the term $J_1$ as 
\begin{equation}
J_1 = \sum_{y} \int_0^T \mathsf E \left[  \frac{\sigma ( y,t) - \sigma( Z_i (t), t) }{ \sigma( Z (t), t) } Q (Z(t),y)  \right] dt  +  \sum_{y} \int_0^T \mathsf E \left[ Q(Z(t),y) \right] dt \, .
\end{equation}
The term in the numerator within the first expectation corresponds to the left hand side of the backward equation obeyed by $\sigma$ (c.f. \eqref{eq:backward_equation}). Inserting this and using $\frac{d}{dx} \log f(x) = \frac{1}{f(x)} \frac{d }{dx} f(x)$ leads to
\begin{equation} \label{eq:J1}
J_1 = -\int_0^T \mathsf E [ \partial_t \log \sigma( Z(t), t) ] dt+ \sum_{y} \int_0^T \mathsf E \left[ Q(Z(t),y) \right]  dt  \, .
\end{equation}
Inserting the relations \eqref{eq:J1} and \eqref{eq:J4} into \eqref{eq:kl_decomposition} and exploiting \eqref{eq:time_dep_backward_equation} we get
\begin{equation*}
 \KL{P^{Z} }{P^{\bar X}} = \KL{P^{Z} }{P^{ X}}  - \int_0^T \frac{d}{dt} \mathsf E[ \log \sigma( Z(t), t)  ]dt \, .
\end{equation*}
Due to the jump conditions of the backward equation, the integral on the right hand side evaluates to
\begin{equation*}
\begin{split}
 \int_0^T \frac{d}{dt} \mathsf E[ \log \sigma( Z(t), t) ] dt  &=  \mathsf E[ \log \sigma( Z(T), T) ]  -  \mathsf E[ \log \sigma( Z(0), 0) ]  \\
 &\quad  + \sum_{k=1}^n \left( \mathsf E[ \log \sigma( Z(t_k^-), t_k^-) ]  -  \mathsf E[ \log \sigma( Z(t_k), t_k) ]  \right)
 \end{split}
\end{equation*}
The terms in the first line correspond to the contributions of the terminal and the initial time and can be ignored. For the terminal time, this follows directly from the terminal constraint $\sigma(x,T) = 1$ for all $x$. The contribution of the initial term, we observe that by definition of $\sigma$ we have
\begin{equation*}
\mathsf E [ \log \sigma( Z(0), 0)] = \mathsf E [ \log p( y_1,\ldots, y_n \mid Z(0) ) ]  = \log p( y_1,\ldots, y_n \mid Z(0) )  =: \log Z \, ,
\end{equation*} 
 which is constant with respect to the variational transition function $\lambda$ since the initial distribution is fixed. Note that in the usual language of variational inference $\log Z$ corresponds to the marginal log likelihood of the data.  Finally, exploiting the reset conditions of the backward equation, we get
\begin{equation*}
 \int_0^T \frac{d}{dt} \mathsf E[ \log \sigma( Z(t), t) ] dt  = \sum_{k=1}^n \mathsf E[ \log p( y_k \mid Z(t_k) ) ] \, .
\end{equation*}
In summary, we get
\begin{equation*}
 \KL{P^{Z} }{P^{\bar X}} = \KL{P^{Z} }{P^{ X}}  - \sum_{k=1}^n \mathsf E[ \log p( y_k \mid Z(t_k) ) ]  + \log Z  \, 
\end{equation*}
which is in line with the usual decomposition of the KL divergence into the evidence $\log Z$ and a free energy contribution $L = \sum_{k=1}^n \mathsf E[ \log p( y_k \mid Z(t_k) ) ] - \KL{P^{Z} }{P^{ X}} $ \cite{blei_2017}.

\subsection{Gradient Descent for MBVI} \label{sec:maximum_principle}

\paragraph{Maximum Principle}
Our goal is to solve the variational problem in the form of the control problem as given in \eqref{eq:control_problem}. However, here we consider a slightly  more general scenario in which we do not solve for the natural moments $\varphi$ directly. Instead, we choose a collection of moment function $\psi$ such that the natural moments can be represented by
\begin{equation}
\varphi(t) = g( \psi (t) )
\end{equation}
for a suitable map $g$. Note that we can always find such a collection of moments by choosing  $\psi = \varphi$ and $g = \mathrm{Id}$. In this formulation, control problem \eqref{eq:control_problem} becomes
\begin{equation} \label{eq:control_problem_mod}
\begin{aligned}
\mathrm{minimize} & &  &L[ \lambda, \psi] - F[\psi] \\
\mathrm{subject \ to} & &  &\frac{d}{dt} \psi(t) = f( \lambda(t), \psi(t) )  ]  \,  \\   
\end{aligned} \, 
\end{equation}
where we assume the initial conditions as fixed and known. We follow the indirect approach known from optimal control and variational calculus by introducing the Langrange multiplier functions (or co-states) $\eta_i$, $i= 1, \ldots, r$ to enforce the ODE relation between $\psi$ and $\lambda$. We obtain the Lagrangian functional
\begin{equation}
J [ \lambda, \psi, \eta] = L[\lambda,\psi] - F[\psi] -\int_0^T \eta(t)^T \left[ f( \lambda (t) , \psi(t) - \dot \psi (t)  \right] \, .
\end{equation} 
where we have stacked the $\eta_i$ into a single vector $\eta$. Since the functional $F$ only acts on the discrete observation times, we will ignore it for a moment. Computing the functional derivative with respect to $\psi$ and setting it to zero leads to the co-state equations
\begin{equation} \label{eq:costate_equation}
\frac{d}{dt} \eta_i (t) =  \sum_{j=1}^r \frac{d g_j }{ d \psi_i}  \left( 1 - \lambda_j(t) + \lambda_j(t) \log \lambda_j (t) \right) -  \sum_{j=1}^r  \frac{d f_j }{ d \psi_i} \eta_j(t) \, 
\end{equation}
valid in between the observations. At the point of the observations, the functional $F$ will induce jump conditions for $\eta$ given by
\begin{equation} \label{eq:costate_reset}
 \lim_{t \nearrow t_k }\eta_i(t) = \eta_i(t_k) + \frac{d}{ d \psi_i(t_k) } \mathsf E[ p(y_k \mid Z(t_k ) )  ]  \, .
\end{equation}
It is therefore crucial that we express the expected log likelihood with respect to the variational process in terms of the moment functions $\psi$. If this is not naturally possible, we may enlarge the space of moment functions suitably. Next, consider the functional derivatives with respect to $\lambda_i$ leading to
\begin{equation} \label{eq:control_condition}
0 = g_i( \psi(t) ) \log \lambda_i (t) - \sum_{j=1}^r \eta_j(t) \frac{ d f_j }{ d \lambda_i } \, .
\end{equation}
The equations \eqref{eq:costate_equation}, \eqref{eq:costate_reset}, \eqref{eq:control_condition} together with the ODE for $\psi$ from a set of necessary conditions for optimal solutions known as Pontryagin's maximum (minimum) principle. Note that since the function $f$ is typically linear in $\lambda$, it may be possible to solve  \eqref{eq:control_condition}  for $\lambda$ and eliminate it in \eqref{eq:costate_equation}. In general, the benefit of this questionable because the resulting ODE for $\eta$ is highly non-linear. 

\paragraph{Gradient Based Optimization} 

A simple approach to solve to obtain a numerical solution from the maximum principle is the forward backward sweep \cite{mcasey_2012}. Here, one starts with an initial guess for $\lambda$ and solves the forward equation. The forward solution is then used to solve \eqref{eq:costate_equation} backward in time. We may then solve \eqref{eq:control_condition} for $\lambda$ to obtain an update given the forward and backward solution. Iterating this procedure may lead to a stationary point of the functional.  \\ In our applications, this procedure turned out to be highly unstable, probably because the updates are too large in the geometry of the probabilistic manifold defined by $\lambda$. We therefore modified the procedure as follows. We keep the forward an backward solution steps, but instead of solving \eqref{eq:control_condition} for $\lambda$, we understand the r.h.s of \eqref{eq:control_condition} as the gradient of the functional $L$ when considered as a function of $\lambda$ alone. More explicitly, we use
\begin{equation} \label{eq:functional_gradient}
(\nabla L[ \lambda])_i = g_i( \psi(t) ) \log \lambda_i (t) - \sum_{j=1}^r \eta_j(t) \frac{ d f_j }{ d \lambda_i } \, .
\end{equation}
 Using \eqref{eq:functional_gradient} allows to apply a gradient descent type algorithm in the scaling factors $\lambda$ by using update steps of the form
\begin{equation} \label{eq:gradient_step}
\lambda^{(n+1)} =\lambda^{(n)} -h \nabla L[ \lambda^{(n)} ]
\end{equation} 
where $h$ is the step size.  An algorithmic description of this procedure in form of pseudo code is given in Alg. \ref{alg:gradient_descent}. \\
%

 \begin{algorithm}[t]
   \caption{Basic Gradient Descent for MBVI}
   \label{alg:gradient_descent}
\begin{algorithmic}[1]
   \STATE {\bfseries Input:} Initial guess for the scaling factors $\lambda^{(0)}(t)$, \\ initial condition $\psi(0)$.
   \REPEAT
   \STATE Given $\lambda^{(n)}(t)$ and $\psi(0)$, compute $\psi^{(n)}(t)$ using \eqref{eq:control_problem_mod}.
   \STATE Given $\lambda^{(n)}(t)$ and $\psi^{(n)}(t)$, compute $\eta^{(n)}(t)$ using \eqref{eq:costate_equation}, \eqref{eq:costate_reset}.
   \STATE Compute current gradient $\nabla L [ \lambda^{(n)} ]$ according to \eqref{eq:functional_gradient}. 
   \STATE Compute $\lambda^{(n+1)}$ from \eqref{eq:gradient_step}.
   \UNTIL{ $ | L [ \lambda^{(n)} ] - L [ \lambda^{(n-1)} ] | < \mathrm{tolerance}$  }
   \STATE {\bfseries Output:} Optimized variational scaling factor $\lambda^*$ .
\end{algorithmic}
\end{algorithm}

\paragraph{Natural Gradient} 

It is well-known that gradient-based algorithms can perform poorly on manifolds. One solution to this is to incorporate the local geometry of the manifold via its metric tensor $G$. Now consider a family of probability distributions $p_\theta$ parametrized by $\theta$. Then the set of all $\theta$ spans a manifold whose geometric structure is given by the Fisher information matrix \cite{amari_1998}. This defines the so called natural gradient
\begin{equation*}
\tilde \nabla p_\theta = G^{-1} \nabla p_\theta \, .
\end{equation*}
Discrete update steps using the natural gradient corresponds to an approximate steepest descent with respect to the local geometry of $p_\theta$. In order to transfer this setting, we exploit a connection between the KL divergence and the fisher information metric
\begin{equation*} 
D [ p_{ \theta} \, || \, p_{  \theta'} ] = \frac{1}{2} (  \theta - \ \theta')^T \ G (\theta) ( \theta - \theta') + o( (  \theta' - \theta)^2 ) \, ,
\end{equation*}
that is the metric $G$ arises from second order expansion of the KL divergence in the parameter of the second argument. While the Fisher information matrix requires a finite dimensional parameter $\theta$, the second order expansion of the KL divergence can be transferred to the path space setting. Thus, consider the variational family corresponding to a partition $\Pi$ of the transition space and consider variational scaling factors $\lambda$, $\lambda'$. Then the divergence of $Z^\lambda $ and $Z^{\lambda'}$ as in Section 3.1 of the main text can be written as
\begin{equation*} \label{eq:vi_family_kl}
D \left[ P^\lambda \, || \, P^{\lambda'} \right] = \int_0^T \varphi_i(t) \left( \lambda'_i(t) - \lambda_i (t) + \lambda_i(t) \frac{\lambda_i(t)}{ \lambda_i' (t) } \right) \mathrm{d} t
\end{equation*}
where the $\varphi_i$ depend on $\lambda$ as they are defined as expected values with respect to $Z^\lambda$. Now rewrite the logarithmic term as
\begin{equation} \label{eq:log_rewrite}
 \log \left( \frac{ \lambda_{i}' (t) }{  \lambda_{i} (t)   }  \right) = \log \left( 1 + \frac{\lambda_i'(t) - \lambda_i(t) }{\lambda_i(t)} \right) \, .
\end{equation}
If $ \sup_{ t \in [0,T ] } | \lambda'(t) - \lambda(t)|$ is small, we may use the standard approximation
 \begin{equation} \label{eq:log_expansion}
\log ( 1+ x) = x - \frac{x^2}{2} + O(x^3) \, .
\end{equation}
Applying \eqref{eq:log_expansion} to \eqref{eq:log_rewrite} and inserting the result into \eqref{eq:vi_family_kl} causes the linear terms to cancel and we are left with
\begin{equation} \label{eq:kl_second_order}
D \left[ P^\lambda \, || \, P^{\lambda'} \right] = \frac{1}{2} \int_0^T \sum_{i=1}^R \frac{ \varphi_i(t) )   }{ \lambda_i(t)  } (\lambda_i'(t) - \lambda_i (t) ) ^2  \mathrm{d} t  + \int_0^T O( (\lambda_i'(t) - \lambda_i (t) ) ^3 ) \mathrm{d} t \, .
\end{equation}
We can understand the above expression as the infinitesimal distance between the path distributions corresponding to the variational parameters $\lambda$ and $\lambda'$ for a fixed partition. From the second order expansion \eqref{eq:kl_second_order}, in combination with \eqref{eq:control_condition}, we get the natural gradient 
\begin{equation} \label{eq:natural_functional_gradient}
 \tilde \nabla L[\lambda] =  \lambda_i(t)  \log \lambda_i (t)   - \frac{\lambda_i(t) }{ \varphi_i (t) } \sum_{j=1}^r \eta_j(t) \frac{ d f_j }{ d \lambda_i } \, .
\end{equation}
where $\varphi = g( \psi )$ and $\eta$ are evaluated for the current value of $\lambda$ via the forward and backward equations. To perform the optimization, we simply have to replace the gradient evaluation in Alg. \ref{alg:gradient_descent} by \eqref{eq:natural_functional_gradient}. Doing so not only increased speed and reliability of the optimization but also led to a visually smoother transition from the prior to the posterior in all considered examples.

\subsection{Parameter Inference}

\paragraph{Statistics for Expectation Maximization}

By the modified definition of the $\varphi_i$ as 
\begin{equation*}
\lambda(x,y,t) = \lambda_i(t) h(x,y) \quad \text{for} \quad (x,y) \in \Pi_i \, ,
\end{equation*}
The functional $L$ changes slightly to
\begin{equation} \label{eq:parameter_vi_functional}
 L[ \varphi, \lambda,\theta]  = \sum_{i=1}^r \int_0^T \varphi_i(t) \left( c_i(\theta) - \lambda_i (t) + \lambda_i(t) \log \frac{ \lambda_i(t) }{ c_i (\theta)}  \right) dt \, .
\end{equation}
By breaking \eqref{eq:parameter_vi_functional} down into individual terms, it becomes clear how the summary statistics
\begin{equation}
\begin{split}
G_i &= \int_0^T \varphi_i(t) dt \, , \\
H_i &= \int_0^T \varphi_i(t) \lambda_i (t) dt \,  
\end{split}
\end{equation}
arise. Considering $L$ as a function of $\theta$ for fixed $\lambda$ and $\varphi$, we may write
\begin{equation} \label{eq:L_of_theta} 
L[ \theta ] = \sum_{i=1}^r  \left( G_i c_i (\theta) - H_i \log c_i (\theta) \right) + const \, . 
\end{equation}
From the last expression, we obtain a stationarity condition by differentiating with respect to $\theta$. 

\paragraph{Bayesian Approach}

Consider a general scenario where with a hierarchical model given by 
\begin{equation}
p(\theta, x, y) = p(\theta) p(x \mid \theta) p( y \mid x) 
\end{equation}
where we aim to approximate the joint posterior $p(\theta, x \mid y)$ by a product $q(\theta) q(x)$. It is straightforward to show that the optimal parameter posterior $q^*(\theta)$ satisfies
\begin{equation}
q^*(\theta) \propto p(c) \exp \left(  -D [ q(x) \, || p(x \mid \theta) p( y \mid x) ]   \right) \, .
\end{equation}
Transferred to our setting, the expression within the exponential becomes the functional $L$ and since we only care about the parts depending on $\theta$, we may as well insert \eqref{eq:L_of_theta} leading to \eqref{eq:parameter_variational_posterior}. 

\section{Examples}

Here we provide explicit forms of the variational functionals of the different examples and the corresponding stationarity conditions.

\subsection{Gene Expression Model}

For the gene expression model, the functions $\varphi_i$ can be expressed in terms of the first order expectations
\begin{align*}
m_i (t) &= \mathsf E[ Z_i(t) ] \, .
\end{align*}
Due to the Gaussian observation model, we also require second order moments
\begin{equation*}
m_{ij} (t) = \mathsf E[ (Z_i(t) Z_j(t)) ]  \, .
\end{equation*}
It is therefore convenient to parametrize the dynamical system in these terms. For the first order moments, we obtain the equations
\begin{align*}
\frac{d}{dt} m_1 (t) = \lambda_1(t)(1-  m_1 (t) ) - \lambda_2(t) m_1 (t)  \, , \\
\frac{d}{dt} m_2(t)  = \lambda_3(t)m_1(t) - \lambda_4(t) m_2(t)  \, , \\
\frac{d}{dt} m_3(t)  \rangle = \lambda_5 m_2(t)  - \lambda_6(t)  m_3(t)  \, .
\end{align*}
For three species, we get 6 additional second order equations
\begin{align*}
\frac{d}{dt} m_{11}(t)  &= \lambda_1 (t) m_1(t)  +\lambda_1(t) - 2 \lambda_1 (t)m_{11}(t) - 2 \lambda_2 m_{11}(t) + \lambda_2(t) m_1(t)  \, , \\
\frac{d}{dt} m_{12} (t) &= \lambda_1(t) m_2(t)  - \lambda_1(t) m_{12} (t)  - \lambda_2(t) m_{12} (t) + \lambda_3(t) m_{11} - \lambda_4 m_{12}(t)  \, ,  \\
\frac{d}{dt} m_{13} (t) &=   \lambda_1(t)m_3(t) - \lambda_1(t) m_{13} (t)  - \lambda_2(t) m_{13}(t)+ \lambda_5(t) m_{12}(t) - \lambda_6(t) m_{13} (t)   \, ,  \\
\frac{d}{dt} m_{22}(t) &=  2 \lambda_3(t) m_{12}(t) + \lambda_3(t) m_1(t)  - 2 \lambda_4(t) m_{22}(t) + \lambda_4(t) m_2(t)  \, ,  \\
\frac{d}{dt} m_{23} (t)  &= \lambda_3(t) m_{13}(t) - \lambda_4(t) m_{23} (t) + \lambda_5(t) m_{22} (t)  \lambda_6(t) m_{23} (t)  \, ,  \\
\frac{d}{dt} m_{33}(t)  &= 2 \lambda_5(t) m_{23} (t) + \lambda_5 (t) m_2(t) - 2 \lambda_6(t) m_{33}(t) + \lambda_6(t) m_3(t)  \, .  \\ 
\end{align*}

\subsection{Predator Prey Model}

As before, we take into account the first and second order moments and choose a corresponding parametrization. Suppressing the explicit time arguments for the sake of readability, the resulting system is given by
\begin{align*}
\dot m_1&= \lambda_1 m_1 - \lambda_2 \left( m_{12} +m_1 m_2  \right) \, , \\
\dot m_2 &= \lambda_3 \left( m_{12} +m_1 m_2  \right) - \lambda_4 m_2  \, , \\
\dot m_{11} &= 2 \lambda_1 m_{11}+\lambda_1 m_1+\lambda_2 (m_{12}+m_1 m_2) + 2 \lambda_2 m_1 \left( m_{12}+m_1 m_2 \right) -2 \lambda_2 m_{112}  \, , \\
\dot m_{12} &=  \lambda_1 m_{12}+\lambda_2 (m_{12}+m_1 m_2) m_2-\lambda_3 (m_{12}+m_1 m_2) m_1-\lambda_4 m_{12} -\lambda_2 m_{122}+\lambda_3 m_{112} \, , \\
\dot m_{22} &= -2 \lambda_3 m_2 (m_{12}+m_1 m_2)+\lambda_3 (m_{12}+m_1 m_2)-2 \lambda_4 m_{22}+\lambda_4 m_2 +  2 \lambda_3 m_{122} \, . 
\end{align*}
As is typical for systems with non-linear intensity functions, the moment equation up to order three are not closed but depend on the (non-central) third-order moments $m_{112}$ and $m_{122}$. In order to obtain a finite dimensional system, we have to use a moment closure method that expresses the higher order moments in terms of lower order moments
\begin{align*}
m_{112} &= V_1( m_1, m_2, m_{11}, m_{12}, m_{22} ) \, , \\
m_{122} &= V_2 ( m_1, m_2, m_{11}, m_{12}, m_{22} ) \, .
\end{align*}
While many standard moment closure approaches use ad hoc choices for the closure functions $V_1, V_2$, we follow the recently proposed variational moment closure approach \cite{bronstein_2018} that allows for a systematic derivation of closure functions based on a set of moment functions and a distributional ansatz. 
In particular we use a log-normal product Poisson mixture distribution and get
\begin{equation}
\mathsf E[ X_1^2 X_2 ]  =  \frac{ \left( \mathsf E[ X_1^2] -\mathsf E[X_1] \right) \mathsf E[X_1 X_2 ]^2  }{\mathsf E[X_1]^2 \mathsf E[X_2] }+\mathsf E[X_1 X_2 ]
\end{equation}
and a similar expression for $\mathsf E[ X_1 X_2^2 ]  $.